\documentclass[runningheads]{llncs}

\usepackage[T1]{fontenc}

\usepackage{graphicx,verbatim}
\usepackage{adjustbox}
\usepackage{float}
\usepackage{multirow}
\usepackage{booktabs}
\usepackage{amsfonts}       
\usepackage{amsmath}
\usepackage{amssymb}
\usepackage{xcolor}
\usepackage{hyperref}           
\usepackage[capitalise]{cleveref}
\usepackage[caption=false,font=footnotesize]{subfig}
\usepackage{microtype}      
\usepackage{placeins}
\usepackage{algorithm}
\usepackage{algorithmic}

\usepackage{color}

\usepackage{graphicx}
\usepackage{hyperref}       
\usepackage{xcolor}

\hypersetup{
    colorlinks,
    linkcolor={red!50!black},
    citecolor={blue!50!black},
    urlcolor={blue!50!black}
}

\begin{document}

\title{Causal Representation Learning with Observational Grouping for CXR Classification}
\titlerunning{Causal Representation Learning with Observational Grouping}
\author{Rajat Rasal\thanks{Corresponding author} \and Avinash Kori \and Ben Glocker}
\authorrunning{Rasal et al.}
\institute{Department of Computing, Imperial College London, London, UK \email{\{rrr2417,a.kori21,b.glocker\}@imperial.ac.uk}}

\maketitle              
\begin{abstract}
Identifiable causal representation learning seeks to uncover the true causal relationships underlying a data generation process. In medical imaging, this presents opportunities to improve the generalisation and robustness of task-specific latent features. This work introduces the concept of grouping observations to learn identifiable representations for disease classification in chest X-rays via an end-to-end framework. Our experiments demonstrate that these causal representations improve performance across multiple classification tasks when grouping is used to enforce invariance with respect to race, sex, and imaging views. 

\end{abstract}

\section{Introduction}
It is well established that discriminative models trained on chest X-rays (CXRs) from specific demographic groups can inadvertently rely on group-specific patterns, such as trends associated with sex, race, or the imaging modality. These patterns often fail to generalise to other populations, potentially amplifying health disparities 
\cite{lotter2024acquisition}. Moreover, it remains unclear whether the presence of group characteristics in learned feature representations are useful for downstream predictive tasks \cite{glocker2023algorithmic}. 

Identifiable and causal representation learning offers a promising solution to address these challenges. Identifiable models provide theoretical guarantees for learning representations that are consistent across training configurations by recovering the true underlying generative structure of the data \cite{yao2024unifying}. 
Causal representation learning (CRL), assuming a task-specific data generation process, focuses on learning representations that follow a causal structure which in turn remain invariant across environments (\emph{i.e., populations, groups})\cite{Bernhard2021towardCRL,yao2024unifying}. Together, identifiability and causal invariances are critical for building generalisable and trustworthy medical AI systems \cite{castro2020causality}.

Early works in representation learning, particularly disentanglement methods \cite{higgins2017betavae}, have shown success in controlled synthetic settings with well-defined sources of variation \cite{locatello2019challenging} and in some medical applications \cite{liu2022learning}. However, these approaches typically fall short of achieving identifiability and often struggle to generalise in real-world medical imaging scenarios. Identifiability has been extensively studied in Independent Component Analysis (ICA) \cite{khemakhem2020variational}, where independent features can be recovered up to scaling and permutation. Yet, when observations undergo non-linear mixing, a standard assumption in real-world medical imaging datasets, the recovery of independent latent variables is fundamentally ill-posed \cite{locatello2019challenging}.

In such real-world datasets, disentangled representations alone often fail to support robust generalisation. Learning the underlying causal structure offers a more promising avenue, as causal representations are invariant across environments, improving both interpretability and predictive reliability \cite{Bernhard2021towardCRL,morioka2023causal}, a necessity in autonomous healthcare systems \cite{Pearl_2009}. 
CRL specifically aims to uncover the dependency structure within the latent space. Recent advances in CRL leverage invariances and data symmetries to infer identifiable representations from observational data \cite{yao2024unifying,khemakhem2020variational,morioka2023causal}, while other approaches rely on interventional \cite{ahuja2022interventional} or counterfactual \cite{brehmer2022weakly} signals to improve identifiability.

In medical imaging, learning identifiable representations offers the potential to improve the generalisation and robustness of task-specific features  \cite{castro2020causality,Bernhard2021towardCRL}. Recent approaches have explored grouping observations as a strategy to learn invariant causal representations \cite{morioka2023causal}. Building on these ideas, we propose an end-to-end framework for disease classification in CXRs that learn identifiable, causal representations invariant to demographic and non-demographic properties. Our main contributions are as follows: 
\begin{enumerate}
    \item We present an end-to-end image classification framework which learns identifiable representations via causal representation learning (\cref{sec:observational_grouping}).
    \item We introduce an observational grouping strategy that enforces invariance to population characteristics, such as race, sex, and imaging view (\cref{alg:data_sampling}).
    \item We demonstrate through qualitative and quantitative experiments that our method improves the generalisation and robustness of disease classification across CXR datasets (\cref{sec:experiments}).
\end{enumerate}

\section{Methods}

\subsection{Background}
Our approach to CRL through observational grouping aligns with and enhances existing mutual information-based contrastive learning methods \cite{tsai2020self,tosh2021contrastive,von2021self,roschewitz2024counterfactual}.
These works show that contrastive learning models observational data under non-IID assumptions to extract invariant causal representations. 
It is important to note, however, that our approach is not contrastive by design; rather, contrastive objectives naturally emerge from the causal assumptions we make \cite{zimmermann2021contrastive}.
In particular, \cite{tosh2021contrastive,von2021self} utilises data augmentations for observational grouping to disentangle invariant content features from style features. 
\cite{roschewitz2024counterfactual} takes an interesting approach, where the notion of grouping is achieved by generating counterfactual pairs based on desired attributes.
In contrast, our work explicitly groups observations based on protected characteristics or other relevant subgroup attributes, thereby inducing non-IID dependencies \cite{morioka2023causal}, to jointly learn attribute-invariant representations and tackle a downstream classification task.

\subsection{Problem Setup}
\label{sec:assumptions}
We consider a dataset that is partitioned into groups based on an observed attribute.
Specifically, the set of groups is defined as $G = \{G_k\}^K_{k=1}$ where each group $G_k$ has $M_k$ samples: $G_k = \{ x^i_k \}^{M_k}_{i=1}$.
We assume that our dataset is non-IID, where, in addition to the imposed grouping structure, each group may follow a different underlying distribution. This can arise, for example, when data is collected from different hospitals, imaging devices, or patient populations.
We define a feature extractor $\phi$ that smoothly maps each image $x$ to a feature representation $z = \phi(x) \in \mathbb{R}^{N}$.
Our classifier is defined as $\Phi = (\psi \circ \phi)$, where $\psi$ is a binary classifier mapping $z$ to class labels $y \in \{0, 1\}$ via a linear layer followed by a sigmoid activation.
Our goal is for representations $z$ to be invariant to differences across the groups in $G$; latent features should not be separable based on the group $G_k$ from which the image is drawn.
Additionally, representations should support accurate classification.
The invariance properties of $\phi$ follow \cite{yao2024unifying}.
Formal definitions are provided in \cref{app:assumptions}. 

\subsection{Invariance with Observational Grouping for Classification}
\label{sec:observational_grouping}

\begin{figure*}[!t]
    \centering
    \includegraphics[width=.95\linewidth]{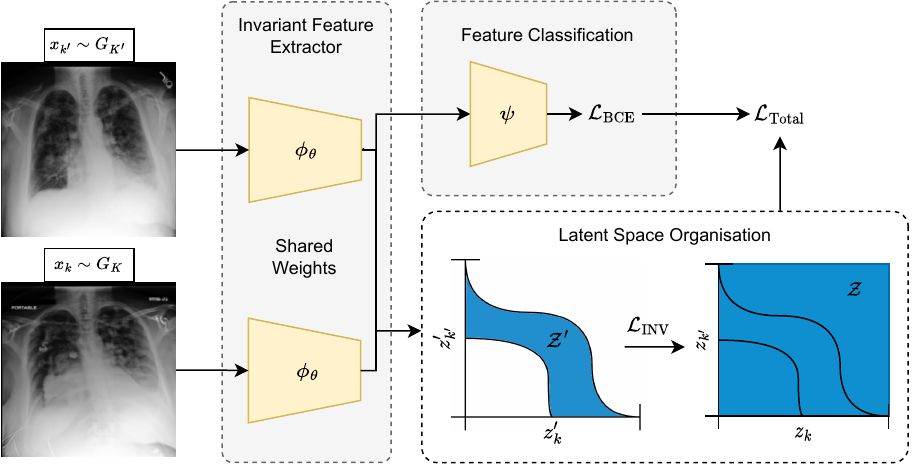}
    \caption{Training with observational grouping organises the latent space such that representations are invariant to properties across groups in $G$. We select two groups randomly $k, k' \sim [K]$ and sample images $x_k \sim G_k$ and $x_{k'} \sim G_{k'}$.
    We use $\phi$ to extract features, and jointly incorporate them into an invariant loss ($\mathcal{L}_{\mathrm{INV}}$) and a binary classification loss ($\mathcal{L}_{\mathrm{BCE}}$). 
    The invariant loss ($\mathcal{L}_{\mathrm{INV}}$) structures the latent space by relaxing the IID assumption, leading to a transformation from \( \mathcal{Z}' \) to \( \mathcal{Z} \), as illustrated. Here, the blue coloured regions indicate the theoretical $\phi$-supported region of the distribution before and after the use of $\mathcal{L}_{\mathrm{INV}}$. The resulting representation is used for classification with $\psi$.}
    \label{fig:latent-org}
\end{figure*}

In the case of CXRs, we define observational groups $G$ based on patient sex, race, or imaging view. Our objective is to learn feature representations that are invariant to these group attributes; in other words, we aim to capture patterns that are shared across all groups while supporting accurate disease classification.
We hypothesise that explicitly leveraging group information leads to better, shared representations for a task. In contrast, giving the model more freedom may allow it to learn group-specific modes in its latent space, limiting generalisation \cite{glocker2023algorithmic}.

\subsubsection{Causal Representations} 
Following \cite{von2021self} and the setup in \cref{sec:assumptions}, a group-invariant feature extractor $\phi$ leveraged for a discriminative task \textit{implicitly} minimises the following objective:
\begin{align}
    \mathcal{L}_{\mathrm{INV}} = \underbrace{\sum_{k < k' \in [1, K]} \mathbb{E} \left[ \| \phi(x_k) - \phi(x_{k'}) \|^2 \right]}_{\text{Similarity}}
    - \underbrace{\sum_{k \in [1, K]} H(\phi(x_k))}_{\text{Uniformity}},
\label{eqn:inv-rep}
\end{align}
where $x_k$ and $x_{k'}$ are sampled from groups $G_k$ and $G_{k'}$. The similarity term encourages representations from different groups to be close, promoting invariance to group-specific variations. The uniformity term maximises entropy of representations, preventing collapse to trivial solutions and ensuring that the learned features remain informative and diverse \cite{yao2023multi}. 
Here, the contrastive-style objectives naturally emerge from the causal assumptions about the data generation process and grouping structure \cite{zimmermann2021contrastive}. Specifically, by treating group membership as soft interventions on latent variables, our method causally identifies invariant content that is shared across groups while discarding group-dependent variations \cite{von2021self}.

\subsubsection{Training Objective} 
In this work, we \textit{directly} optimise $\mathcal{L}_{\mathrm{INV}}$  alongside the cross-entropy loss $\mathcal{L}_{\mathrm{BCE}}$ to jointly learn the classifier $\psi$ and invariant feature extractor $\phi$:
\begin{align}
\mathcal{L}_{\mathrm{Total}} = \mathcal{L}_{\mathrm{INV}} + \mathcal{L}_{\mathrm{BCE}}.
\label{eq:l_total}
\end{align}
The end-to-end invariant training procedure is illustrated in \cref{fig:latent-org}, where we sample images $x_k$ and $x_{k'}$ from groups $G_k$ and $G_{k'}$, corresponding to \texttt{male} and \texttt{female} patients, respectively, both with $\texttt{pleural effusion}$. $\mathcal{L}_{\mathrm{INV}}$ organises the latent space to capture group-invariant features responsible for \texttt{pleural effusion}. Intuitively, minimising $\mathcal{L}_{\mathrm{INV}}$ reduces latent intra-class spread across groups, while $\mathcal{L}_{\mathrm{BCE}}$ improves inter-class separation ($\delta$). We hypothesise that this combination not only improves predictive performance but also reduces variance in downstream prediction. We also provide pseudocode for training in \cref{alg:data_sampling}.

\subsubsection{Identifiability}
Following the setup in \cref{sec:assumptions}, we can apply the identifiability result from \cite{von2021self}. This guarantees that the invariant content variable $z = \phi(x_k) = \phi(x_{k'})$ is uniquely identified by $\phi$. This demonstrates theoretically that the learned representations are stable and reproducible across training configurations, and are robust to distribution shift.

\subsection{Data and Implementation Details}

\subsubsection{Datasets}
We use the \textsc{CheXpert} \cite{irvin2019chexpert} and \textsc{MIMIC} \cite{johnson2016mimic,johnson2019mimic} datasets in this study. We extract a subset containing only images labelled as \texttt{no findings}, \texttt{pleural effusion} or \texttt{cardiomegaly}, with dataset splits included in
\cref{app:dataset_info}.
All images are resized to a resolution of $224 \times 224$. \cref{alg:data_sampling} details the sampling process used during invariant training. In contrast, the validation and test sampling procedures are the same as the baseline classifiers.

\begin{algorithm}[!t]
  \caption{Data sampling and forward pass for training invariant model.}
  \label{alg:data_sampling}
  \begin{algorithmic}[1]
    \renewcommand{\baselinestretch}{1.2}\selectfont
    \STATE \textbf{Input:} Groups $G = \{G_k\}^K_{k=1}$; Invariant feature extractor $\phi$; Samples per iteration $P \geq 2$; Dataset classes $\mathcal{Y}$. \hfill {\color{gray}$\triangleright$ Groups and labels}   
    \STATE \textbf{Sampling:}
    \STATE \hspace{1cm} Select class $y \sim \mathcal{U}(\mathcal{Y})$ \hfill {\color{gray}$\triangleright$ Disease selection, $\mathcal{U}$ uniform distribution}
    \STATE \hspace{1cm} $\mathcal{G} = \{ G_p \sim \mathcal{U}(G) \mid p \in [1, P] \}$   \hfill {\color{gray}$\triangleright$ Subgroups selection}
    \STATE \hspace{1cm} $\mathcal{X} = \{ x_p \sim \mathcal{U}(G_p \mid y) \mid G_p \in \mathcal{G} \}$ \hfill {\color{gray}$\triangleright$ Sampled subgroups filtered by $y$}

    \STATE \textbf{Forward Pass:}
    \STATE \hspace{1cm} 
    $\mathcal{L}_{\mathrm{INV}} = 0, \mathcal{L}_{\mathrm{BCE}} = 0, x_{k'} \sim \mathcal{X}$ \hfill {\color{gray}$\triangleright$ Init. losses and select reference image.}
    \STATE \hspace{1cm} \textbf{for} $x_k \in \mathcal{X}$:
    \STATE \hspace{1.5cm} $z_k \leftarrow \phi(x_k), z_{k'} \leftarrow \phi(x_{k'})$ \hfill {\color{gray}$\triangleright$ Extract invariant features}
    \STATE \hspace{1.5cm} $\mathcal{L}_{\mathrm{INV}} \leftarrow \mathcal{L}_{\mathrm{INV}} + \| z_k - z_{k'} \|^2 + H(z_k)$ \hfill {\color{gray}$\triangleright$ Invariance loss}
    \STATE \hspace{1.5cm} $\mathcal{L}_{\mathrm{BCE}} \leftarrow \mathcal{L}_{\mathrm{BCE}} + \sum \text{BCEWithLogits}(y, \psi(z_k))$ \hfill {\color{gray}$\triangleright$ Classification loss}
    \STATE \hspace{1.5cm} $x_{k'} \leftarrow x_k$

    \STATE \hspace{1cm} $\mathcal{L}_{\mathrm{Total}} = \mathcal{L}_{\mathrm{INV}} + \mathcal{L}_{\mathrm{BCE}}$ \hfill {\color{gray}$\triangleright$ Total loss}
  \end{algorithmic}
\end{algorithm}

\subsubsection{Models and Metrics}
Our invariant representation learning strategy is assessed across three deep learning backbones implementing $\phi$: ResNet-18 \cite{he2016deep}, DenseNet-121 \cite{huang2017densely} and EfficientNetB0 \cite{tan2019efficientnet}. These models were selected for their strong performance in prior works on CXR analysis \cite{sellergren2022simplified,rajpurkar2017chexnet}. By applying our strategy across diverse architectures, we ensure robustness in learning disease-specific features while remaining invariant to confounder such as imaging view (\texttt{AP, PA}), sex (\texttt{male, female}), and race (\texttt{white, black, asian}). We use the area under the receiver operator curve (AUROC) to compare models trained conventionally with $\mathcal{L}_{\mathrm{BCE}}$ against models trained with $\mathcal{L}_{\mathrm{Total}}$ via the invariant strategy. 

\subsubsection{Training}
We train invariant models with two loss components, as seen in \cref{eq:l_total}; the binary cross-entropy loss $\mathcal{L}_{\mathrm{BCE}}$ and the proposed invariance loss $\mathcal{L}_{\mathrm{INV}}$. 
We ensure that the gradients from $\mathcal{L}_{\mathrm{INV}}$ update the parameters of $\phi$, while $\mathcal{L}_{\mathrm{BCE}}$ updates the parameters of $\psi$.
We apply the same set of hyper-parameters for all models in all our experiments: batch size of $32$, learning rate of $0.001$ for parameters in $\phi$, and learning rate of $0.0001$ for parameters in $\psi$. We learn all models using Adam optimisation for $20$ epochs, selecting models with the largest $\mathcal{L}_\mathrm{BCE}$ on the validation set for analysis in \cref{sec:experiments}.

\section{Experiments}
\label{sec:experiments}

\subsection{Identifiability Analysis}
\label{sec:identifiability_analysis}
We first analyse the latent representations of our model to explicitly evaluate identifiability. For this, we learn every $\phi$ and dataset combination over 5 random seeds and measure the Mean Correlation Coefficient (MCC) \cite{khemakhem2020ice} over 1000 latent samples. Across all considered invariances and disease classes, we observe an MCC of around $\mathbf{0.99}$ with \textsc{ResNet}, $\mathbf{1.0}$ with \textsc{EfficientNet}, and $\mathbf{0.99}$ with \textsc{DenseNet}. 
The high MCC values (near or equal to 1.0) across experiments indicate model convergence in similar local minima irrespective of the starting parameters dictated by random seeds during initialisation. 
This shows that our representations are stable and unique, crucial for robustness in medical imaging.

\subsection{Impact of Invariance Loss}

\begin{table}[t]
\centering
\caption{AUROC for invariant and non-invariant classifiers.}
\resizebox{\textwidth}{!}{
\begin{tabular}{l|c|c|c|c}
\hline
& \multicolumn{1}{c}{\,{\scriptsize \textbf{CheXpert}}\,} 
& \multicolumn{1}{|c}{\,{\scriptsize \textbf{MIMIC}}\,} & \multicolumn{1}{|c}{\,{\scriptsize \textbf{CheXpert $\rightarrow$ MIMIC}}\,} 
& \multicolumn{1}{|c}{\,{\scriptsize \textbf{MIMIC $\rightarrow$ CheXpert}}\,} \\ 
\hline
& \multicolumn{4}{c}{\texttt{no findings} vs \texttt{pleural effusion}} \\
\hline
\textsc{ResNet} & 93.04 ± 0.88 & 94.31 ± 0.11 & 90.87 ± 1.51 & 94.08 ± 0.32 \\
$\sim$ \textsc{View Inv.} & 93.78 ± 0.21 & 93.87 ± 0.32 & 92.60 ± 0.52 & 93.78 ± 0.38 \\
$\sim$ \textsc{Race Inv.} & 94.23 ± 0.09 & 94.04 ± 0.08 & 92.78 ± 0.10 & 93.97 ± 0.23 \\
$\sim$ \textsc{Sex Inv.} & 94.46 ± 0.13 & 94.30 ± 0.14 & 93.27 ± 0.29 & 94.28 ± 0.09 \\
\hline
\textsc{EfficientNet} & 92.78 ± 0.43 & 94.62 ± 0.10 & 91.27 ± 0.85 & 94.51 ± 0.15 \\
$\sim$ \textsc{View Inv.} & 93.46 ± 0.47 & 94.20 ± 0.12 & 91.95 ± 1.19 & 93.88 ± 0.28 \\
$\sim$ \textsc{Race Inv.} & 94.32 ± 0.10 & 94.25 ± 0.10 & 93.22 ± 0.14 & 94.15 ± 0.15 \\
$\sim$ \textsc{Sex Inv.} & 94.59 ± 0.15 & 94.46 ± 0.04 & 93.43 ± 0.24 & 94.41 ± 0.18 \\
\hline
\textsc{DenseNet} & 92.78 ± 1.11 & 93.88 ± 0.73 & 90.55 ± 1.53 & 93.81 ± 1.19 \\
$\sim$ \textsc{View Inv.} & 94.22 ± 0.11 & 93.94 ± 0.17 & 93.24 ± 0.26 & 93.88 ± 0.18 \\
$\sim$ \textsc{Race Inv.} & 94.13 ± 0.40 & 93.94 ± 0.14 & 92.69 ± 0.42 & 93.91 ± 0.20 \\
$\sim$ \textsc{Sex Inv.} & 94.31 ± 0.22 & 94.40 ± 0.12 & 93.13 ± 0.34 & 94.32 ± 0.18 \\
\hline
& \multicolumn{4}{c}{\texttt{no findings} vs \texttt{cardiomegaly}} \\
\hline
\textsc{ResNet} & 89.24 ± 1.07 & 87.27 ± 1.03 & 87.35 ± 2.40 & 85.67 ± 2.03 \\
$\sim$ \textsc{View Inv.} & 90.56 ± 0.33 & 91.47 ± 0.49 & 90.24 ± 0.99 & 90.02 ± 1.10 \\
$\sim$ \textsc{Race Inv.} & 90.76 ± 0.28 & 91.59 ± 0.25 & 90.49 ± 0.39 & 89.48 ± 0.89 \\
$\sim$ \textsc{Sex Inv.} & 91.08 ± 0.43 & 92.40 ± 0.07 & 90.71 ± 0.55 & 91.16 ± 0.11 \\
\hline
\textsc{EfficientNet} & 89.24 ± 1.05 & 88.23 ± 1.91 & 87.75 ± 2.23 & 85.91 ± 2.26 \\
$\sim$ \textsc{View Inv.} & 91.07 ± 0.30 & 91.89 ± 0.10 & 90.72 ± 0.52 & 90.28 ± 0.77 \\
$\sim$ \textsc{Race Inv.} & 91.27 ± 0.16 & 92.08 ± 0.10 & 91.03 ± 0.46 & 90.19 ± 0.61 \\
$\sim$ \textsc{Sex Inv.} & 90.61 ± 0.39 & 92.41 ± 0.12 & 89.74 ± 1.24 & 91.18 ± 0.21 \\
\hline
\textsc{DenseNet} & 88.41 ± 0.94 & 87.06 ± 1.42 & 87.43 ± 1.58 & 85.21 ± 1.67 \\
$\sim$ \textsc{View Inv.} & 90.19 ± 1.42 & 91.00 ± 0.47 & 90.75 ± 0.66 & 90.18 ± 0.40 \\
$\sim$ \textsc{Race Inv.} & 90.10 ± 1.54 & 91.60 ± 0.44 & 89.64 ± 1.46 & 89.71 ± 1.38 \\
$\sim$ \textsc{Sex Inv.} & 91.24 ± 0.21 & 92.22 ± 0.12 & 91.35 ± 0.29 & 91.08 ± 0.18 \\
\hline
\end{tabular}
}
\label{table:results}
\end{table}

We evaluate the proposed invariance method across three different architectures for $\phi$ and present results both within and across datasets to assess generalisability.
For this analysis, we train invariant and non-invariant models using 10 different random seeds. We select the top 5 models based on validation AUROC and report the mean and standard deviation of their performance. 
The final tabulated results, summarised in Table \ref{table:results}, provide a comprehensive comparison of the models under different settings. We observe that models trained using the invariance strategy (denoted by the $\sim$ prefix) consistently outperform their respective baseline models, which are listed in the top row of each section. The standard deviation of latent representations in models trained with the invariance strategy is consistently lower than the corresponding baseline, as discussed in \cref{sec:observational_grouping} and examined further in \cref{sec:invariant_representations}. This supports the hypothesis that invariant training leads to more stable convergence, as suggested by our MCC results in \cref{sec:identifiability_analysis}. The impact of group sample sizes on model performance is evident, particularly in the case of sex-based invariance, where the larger sample size in each subgroup appears to contribute to higher overall performance.

\begin{table}[t]
\centering
\setlength{\tabcolsep}{5pt}
\caption{Inter-class separation ($\delta$) for invariant and non-invariant classifiers.}
\begin{tabular}{l|c|c|c|c}
\hline
& \multicolumn{4}{c}{\textbf{CheXpert}} \\
\hline
& \textsc{Non Inv.} & \textsc{View Inv.} & \textsc{Race Inv.} & \textsc{Sex Inv.} \\
\hline
& \multicolumn{4}{c}{\texttt{no findings} vs \texttt{pleural effusion}} \\
\hline
\textsc{ResNet18} & 0.245 ± 0.039 & 0.350 ± 0.011 & 0.406 ± 0.013 & 0.404 ± 0.017 \\
\textsc{EfficientNet} & 0.379 ± 0.047 & 0.353 ± 0.037 & 0.392 ± 0.017 & 0.414 ± 0.035 \\
\textsc{DenseNet} & 0.244 ± 0.017 & 0.330 ± 0.019 & 0.339 ± 0.028 & 0.365 ± 0.025 \\
\hline
& \multicolumn{4}{c}{\texttt{no findings} vs \texttt{cardiomegaly}} \\
\hline
\textsc{ResNet18} & 0.086 ± 0.017 & 0.271 ± 0.024 & 0.314 ± 0.013 & 0.295 ± 0.019 \\
\textsc{EfficientNet} & 0.240 ± 0.029 & 0.293 ± 0.026 & 0.334 ± 0.011 & 0.298 ± 0.021 \\
\textsc{DenseNet} & 0.034 ± 0.011 & 0.248 ± 0.019 & 0.267 ± 0.012 & 0.277 ± 0.012 \\
\hline
& \multicolumn{4}{c}{\textbf{MIMIC}} \\
\hline
& \multicolumn{4}{c}{\texttt{no findings} vs \texttt{pleural effusion}} \\
\hline
\textsc{ResNet18} & 0.281 ± 0.032 & 0.331 ± 0.012 & 0.381 ± 0.018 & 0.377 ± 0.022 \\
\textsc{EfficientNet} & 0.334 ± 0.011 & 0.340 ± 0.027 & 0.389 ± 0.018 & 0.41 ± 0.015 \\
\textsc{DenseNet} & 0.121 ± 0.063 & 0.309 ± 0.021 & 0.342 ± 0.020 & 0.332 ± 0.016 \\
\hline
& \multicolumn{4}{c}{\texttt{no findings} vs \texttt{cardiomegaly}} \\
\hline
\textsc{ResNet18} & 0.137 ± 0.036 & 0.297 ± 0.010 & 0.307 ± 0.018 & 0.318 ± 0.015 \\
\textsc{EfficientNet} & 0.254 ± 0.056 & 0.294 ± 0.006 & 0.339 ± 0.019 & 0.343 ± 0.015 \\
\textsc{DenseNet} & 0.020 ± 0.012 & 0.248 ± 0.015 & 0.269 ± 0.013 & 0.276 ± 0.025 \\
\hline
\end{tabular}
\label{table:delta_results}
\end{table}

\subsection{Analysis of Invariant Latent Representations}
\label{sec:invariant_representations}
\begin{figure}[!t]
  \centering
  \subfloat[Conventionally trained classifier. Density estimation is performed for disease, view, race and sex (left to right).]{%
    \includegraphics[width=\textwidth]{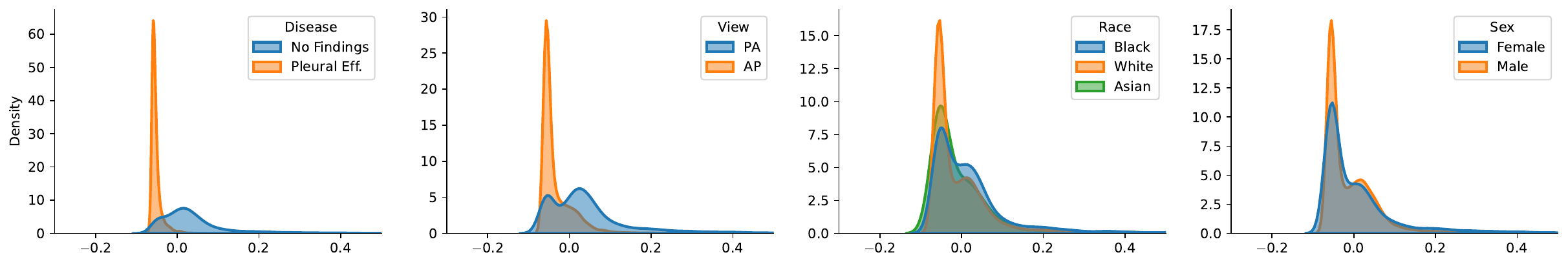}%
    \label{fig:subfig_non_invariance}%
  }
  \hfill
  \subfloat[Invariantly trained classifier w.r.t view, race, and sex invariance (left to right). Density estimation is performed for disease (top) and invariant attributes (bottom).]{%
    \includegraphics[width=\textwidth]{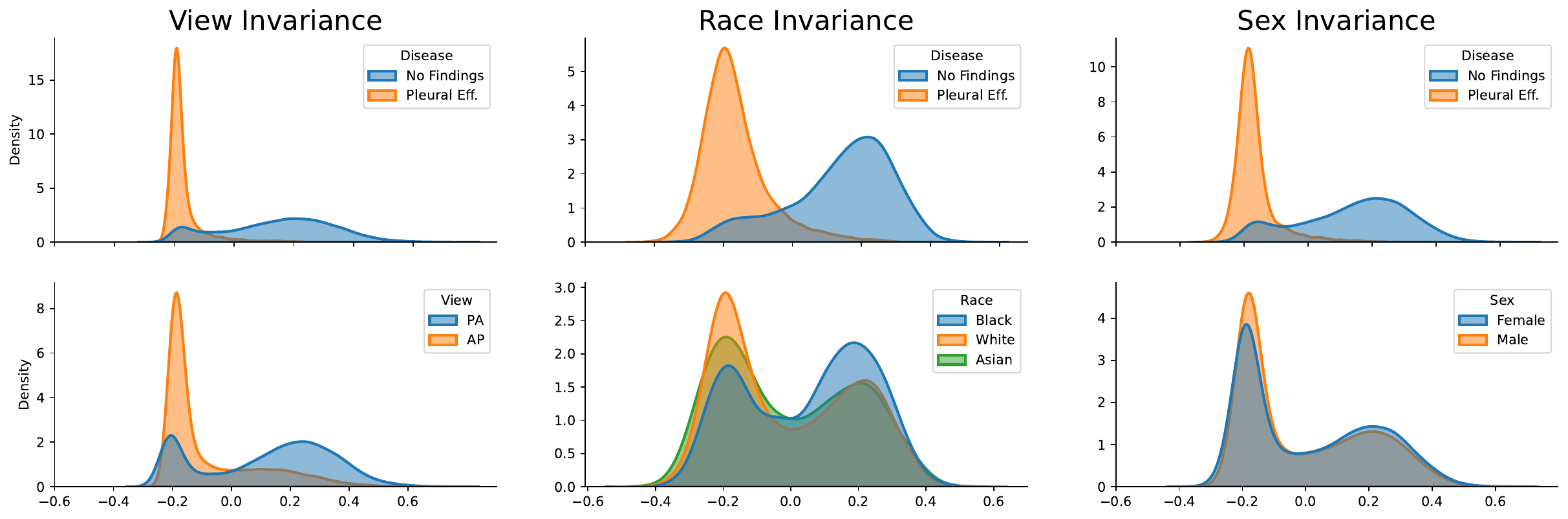}%
    \label{fig:subfig_invariance}%
  }
  \caption[tsne plots]{Density estimation on scaled PC1 embeddings of invariant and non-invariant latent features from $\phi$, implemented with a \textsc{DenseNet} backbone, for \texttt{no findings} vs \texttt{pleural effusion} classification.}
  \label{fig:combined_pca_plots}
\end{figure}

In this section, we analyse the properties of latent representations in models trained conventionally compared to those trained with the proposed invariance strategy. We evaluate this both qualitatively, using principal component analysis (PCA), and quantitatively, using the inter-class separation metric $\delta = \| c_{\mathrm{NF}} - c_{\mathrm{D}} \|^2_2 / s$, where $c$ are first principal component (PC1) of the latent features extracted using $\phi$, $c_{\mathrm{NF}}$ and $c_{\mathrm{D}}$ are the centroids of the \texttt{no findings} and disease class clusters, respectively, and $s = \| c_{\mathrm{MAX}} - c_{\mathrm{MIN}} \|^2_2$ scales the distance. \cref{table:delta_results} shows that inter-class separation ($\delta$) is consistently higher for invariant models, and the standard deviation is consistently lower, indicating improved identifiability in line with the MCC metrics in \cref{sec:identifiability_analysis}. These results provide some intuition for the standard deviations of invariant models being generally lower than those trained conventionally in \cref{table:results}. This is visualised in \cref{fig:subfig_invariance}, which shows the improved linear separability of disease in PC1 with invariant training when compared to PC1 of the conventionally trained model in \cref{fig:subfig_non_invariance}.

\section{Discussion and Conclusion}
In this work, we investigated grouping-based causal representation learning for disease classification in chest X-rays. By leveraging natural groupings in the data, our method improves the linear separability of disease features, leading to better predictive performance and lower error variability. We support these findings through latent space analysis and by observing consistent convergence behaviours across experiments.
Our results, validated on multiple architectures and datasets, show that using our end-to-end invariant training strategy improves model stability and generalisation through identifiability. The choice of grouping attributes was guided by prior research on relevant demographic and non-demographic factors \cite{gichoya2022ai,yang2024limits}.
Future work could explore applying this approach to other medical imaging tasks (e.g. scanner invariance in breast imaging), multi-modal data (e.g. text-image pairs), and alternative grouping criteria (e.g. invariance to multiple groups based on assumptions from a causal graph). 
More broadly, grouping-based methods like ours can be interpreted as enforcing causal invariances to remove spurious shortcuts in classifiers \cite{NEURIPS2022_d791394d}. Enforcing invariances may reduce accuracy or suppress meaningful group differences \cite{zhao2022fundamental,petersen2023demographically}, thereby compromising group-fairness. We therefore advise that how and when to use invariance strategies should be guided by domain-specific fairness analyses and expert knowledge of the causal data-generating process. Our code is available at \url{https://github.com/RajatRasal/CRL-for-CXR-Classification}.

\begin{credits}
\subsubsection{\ackname} We thank Raghav Mehta, Fabio De Sousa Ribeiro, Pavithra Manoj and Fiona Kekwick for their detailed discussions and insightful feedback on early versions of this manuscript. R.R. is supported by the Engineering and Physical Sciences Research Council (EPSRC) through a Doctoral Training Partnerships PhD Scholarship. A.K. was supported by UKRI (grant no. EP/S023356/1), as part of the UKRI Centre for Doctoral Training in Safe and Trusted AI, and acknowledges support from the EPSRC Doctoral Prize. B.G. received support from the Royal Academy of Engineering as part of his Kheiron/RAEng Research Chair and acknowledges the support of the UKRI AI programme, and the EPSRC, for CHAI - EPSRC Causality in Healthcare AI Hub (grant no. EP/Y028856/1).

\subsubsection{\discintname}
B.G. is a part-time employee of DeepHealth. No other competing interests.
\end{credits}

\bibliographystyle{splncs04}
\bibliography{Paper-0024}

\newpage

\appendix

\section{Formal Definitions}
\label{app:assumptions}

\begin{definition}(Invariant Feature Extractor) An $N$-dimensional invariant representation exists for any non-IID dataset $\{x_k \mid x_k \in G_k, \forall k \in [K] \}$. The feature extractor $\phi$ is a smooth function that maps observations to an $N$-dimensional unit cube.  
\end{definition} 

\begin{definition}(Invariance)  
Let $A \subseteq [N]$ be an index subset of $\mathbb{R}^N$ and $\sim_{\iota}$ be an equivalence relation on $ \mathbb{R}^{|A|} $, with the quotient space $M := \mathbb{R}^{|A|} / \sim_{\iota}$ and the projection map $\iota: \mathbb{R}^{|A|} \to M$. Invariance on $A$ is satisfied when two vectors $a, b \in \mathbb{R}^{|A|}$, defined as $a = \phi(x^i_k), \; b = \phi(x^j_k)$ with $i,j \in [1,K]$, are \textbf{invariant under} $\iota$, hence belonging to the same equivalence class: $\iota(a) = \iota(b) \iff a \sim_{\iota} b$.  
\label{ass:invariance}  
    
\end{definition}

\begin{definition}(Identifiability)
    Given $\theta$ as the set of parameters of the invariant feature extractor, the equivalence relation $\sim_{\iota}$ on $\theta$ is defined as:
\begin{align}
    &\theta \sim_{\iota} \tilde{\theta} \iff \exists \; \mathbf{A} \in \mathbb{R}^{|A| \times |A|}, \; \mathbf{b} \in \mathbb{R}^{|A|} \nonumber \\
    \text{s.t.} \; \; & \phi_{\theta}(x) = \mathbf{A} \phi_{\tilde{\theta}}(x) + \mathbf{b}, \; \forall x \in \bigcup_{k\in[1,K]} G_k,
\label{eqn:Dequivalence}
\end{align}
where $\mathbf{A}$ is an affine transformation matrix and $\mathbf{b}$ is a translation vector.
\label{dfn:e-equivalence}
    
\end{definition}

\begin{definition}(Invariant Representations)
Consider a dataset $\mathcal{X} = \{x_k \mid x_k \sim G_k, \forall k \in [1, K]\}$ and an invariant feature extractor $\phi$. Minimising the following objective results in representations invariant to groups in $G$:
\begin{align}
    \mathcal{L}_{\mathrm{INV}} = \sum_{k < k' \in [1, K]} \mathbb{E} \left[ \| \phi(x_k) - \phi(x_{k'}) \|^2 \right]
    - \sum_{k \in [1, K]} H(\phi(x_k)),
\end{align}
where expectation is with respect to $p(\mathcal{X})$ and \( H(\cdot) \) denotes the differential entropy. 
    
\end{definition} 

\section{Dataset Info}
\label{app:dataset_info}

\begin{table}[h]
\centering
\caption{Data splits for MIMIC and CheXpert}
\begin{tabular}{l|c|c|c|c|c|c}
\hline
& \multicolumn{3}{c|}{\textbf{CheXpert}} & \multicolumn{3}{c}{\textbf{MIMIC}} \\ 
\hline
& \multicolumn{1}{c|}{\,\textbf{Train} \,} & \multicolumn{1}{c|}{\, \textbf{Val} \,} & \multicolumn{1}{c|}{\, \textbf{Test} \,} & \multicolumn{1}{c|}{\, \textbf{Train} \,} & \multicolumn{1}{c|}{\, \textbf{Val} \,} & \multicolumn{1}{c}{\, \textbf{Test} \,} \\ 
\hline
\texttt{no findings} & 6,514 & 1,086 & 3,316 & 34,530 & 5,393 & 16,692 \\
\texttt{pleural effusion} & 31,015 & 5,049 & 15,510 & 27,806 & 4,575 & 13,843 \\
\texttt{cardiomegaly} & 9,353 & 1,548 & 4,889 & 22,157 & 3,660 & 11,484 \\
\hline
\end{tabular}
\label{table:splits}
\end{table}

\end{document}